# Multi-SIGATnet: A multimodal schizophrenia MRI classification algorithm using sparse interaction mechanisms and graph attention networks

Yuhong Jiao, Jiaqing Miao, Jinnan Gong, Hui He, Ping Liang, Cheng Luo and Ying Tan, *Member, IEEE*

**Abstract**—Schizophrenia is a serious psychiatric disorder. Its pathogenesis is not completely clear, making it difficult to treat patients precisely. Because of the complicated non-Euclidean network structure of the human brain, learning critical information from brain networks remains difficult. To effectively capture the topological information of brain neural networks, a novel multimodal graph attention network based on sparse interaction mechanism (Multi-SIGATnet) was proposed for SZ classification was proposed for SZ classification. Firstly, structural and functional information were fused into multimodal data to obtain more comprehensive and abundant features for patients with SZ. Subsequently, a sparse interaction mechanism was proposed to effectively extract salient features and enhance the feature representation capability. By enhancing the strong connections and weakening the weak connections between feature information based on an asymmetric convolutional network, high-order interactive features were captured. Moreover, sparse learning strategies were designed to filter out redundant connections to improve model performance. Finally, local and global features were updated in accordance with the topological features and connection weight constraints of the higher-order brain network, the features being projected to the classification target space for disorder classification. The effectiveness of the model is verified on the Center for Biomedical Research Excellence (COBRE) and University of California Los Angeles (UCLA) datasets, achieving 81.9\% and 75.8\% average accuracy, respectively, 4.6\% and 5.5\% higher than the graph attention network (GAT) method. Experiments showed that the Multi-SIGATnet method exhibited good performance in identifying SZ.

**Index Terms**—multimodality; graph convolution network; computer-aided diagnosis; schizophrenia; classification

## I. Introduction

Schizophrenia (SZ), an enigmatic and chronic psychiatric disorder, predominantly emerges during the transitional phase of late adolescence or early adulthood, exerting a profound impact on affected individuals. The disorder is pervasive, afflicting an estimated 1% of the global population [1]. Despite its prevalence, the etiology of SZ remains shrouded in ambiguity, implicating a complex interplay of genetic, environmental, and neurobiological factors that elude straightforward elucidation. The diagnostic process of SZ is notably challenging due to the absence of concrete, objective criteria. Clinical diagnoses are predominantly anchored in the subjective assessments made by mental health professionals, often supplemented by the application of various symptom-rating scales [2]. This reliance on subjective evaluation renders the diagnostic process susceptible to variability in interpretation, thereby increasing the risk of both underdiagnosis and misdiagnosis. The heterogeneity of symptom presentation further compounds the intricacy of early SZ detection, rendering its timely diagnosis an onerous and unresolved endeavor in the field of psychiatry [3].

Magnetic resonance imaging (MRI) technology, a radiation-free method of brain imaging, provides new perspectives for the objective analysis of schizophrenia (SZ). Structural MRI (sMRI), with its superior spatial resolution, offers precise visualization of brain anatomy and reveals subtle features of tissue structure [4, 5]. Functional MRI (fMRI), on the other hand, reflects changes in the local activity of brain tissues, providing a dynamic perspective for understand-ing brain function [6–8]. With the aid of multimodal data fusion, the integration of information from different imaging modalities can effectively enhance feature representation

Yuhong Jiao is with the Key Laboratory for Computer Systems of State Ethnic Affairs Commission, Southwest Minzu University, Chengdu 610041, China and is with School of computing and software,Chengdu Neusoft University, Chengdu 611844, China .E-mail: JiaoYuhong@nsu.edu.cn.

Jiaqing Miao and Ping Liang are with the Key Laboratory for Computer Systems of State Ethnic Affairs Commission, Southwest Minzu University, Chengdu 610041, China. E-mail: liang_p@hotmail.com, mjq_011114117@163.com.

Jinnan Gong is with School of Computer Science, Chengdu University of Information Technology, Chengdu 610225, China and the Key Laboratory for NeuroInformation of Ministry of Education, High-Field Magnetic Resonance Brain Imaging Key Laboratory of Sichuan Province, Center for Information in Medicine, University of Electronic Science and Technology of China, Chengdu 610054, China. E-mail: 18680807330@163.com.

Hui He is the Key Laboratory for NeuroInformation of Ministry of Education, High-Field Magnetic Resonance Brain Imaging Key Laboratory of Sichuan Province, Center for Information in Medicine, Uni-versity of Electronic Science and Technology of China, Chengdu 610054, China. E-mail: hehui_hhwdwd@163.com.

Corresponding author: Ying Tan is with the Key Laboratory for Computer Systems of State Ethnic Affairs Commission, Southwest Minzu University, Chengdu 610041, China. E-mail: ty7499@swun.edu.cn.

Cheng Luo is with the Key Laboratory for NeuroInformation of Ministry of Education, High-Field Magnetic Resonance Brain Imaging Key Laboratory of Sichuan Province, Center for Information in Medicine, University of Elec-tronic Science and Technology of China, Chengdu 610054, China. E-mail: chengluo@uestc.edu.cn.

capabilities and improve model predictive performance. In the diagnosis and classification of schizophrenia, multimodal data fusion techniques have demonstrated their unique value [9]. For instance, by combining fMRI, sMRI, and genetic information, researchers can delve deeper into exploring the underlying neurobiological mechanisms of SZ. Studies by [10] Masoudi et al. [10] and Li et al. [11] have shown that the use of canonical correlation analysis in conjunction with a deep learning model can effectively integrate information from different modalities, thereby improving the classification accuracy of SZ. Furthermore, Luo et al. [12] employed three-way parallel independent component analysis to jointly analyze fMRI, sMRI, and single nucleotide polymorphism data, uncovering potential genetic- brain-cognitive regulatory pathways in SZ. However, the completeness of multi-modal data is essential for accurate diagnosis. Abdelaziz et al. [13] addressed the issue of missing sample information by proposing a solution based on linear interpolation. This method enhances the classification accuracy of Alzheimer's disease by synthesizing missing features of samples and merging multimodal neuroimaging with genomic data. The application of this approach offers a novel strategy for handling incomplete multimodal data. Functional connectivity analysis, an important tool for studying interactions between brain regions, offers new insights for the adjunctive diagnosis of SZ. Abnormal brain connectivity is considered one of the key factors in the pathogenesis of SZ [14]. Therefore, incorporating brain connectivity in the process of multimodal data fusion is vital for a comprehensive understanding of SZ's pathological mechanisms. By combining multimodal MRI techniques with functional connectivity analysis, researchers can assess the brain's functional status in schizophrenia patients more thoroughly, offering a scientific foundation for early diagnosis and treatment.

To alleviate the lack of brain topological attributes, a novel multimodal data fusion method based on graph neural networks (GNNs) is proposed. GNNs have made promising progress in brain disease classification by clearly capturing the internal dependencies of brain networks. Using data from demographic studies and neuroimaging, Lie et al. [15] and Zeng et al. [16] used a graph convolutional network (GCN) to predict mild cognitive impairment (MCI). Yao et al. [17] created a multi-scale brain connection network to overcome the constraints of a single brain template by segmenting the brain region template from coarse to fine. They then proposed a multiscale triplet GCN for the identification of attention deficit hyperactivity disorder and MCI, resulting in improved classification performance. Using a graph attention network (GAT) as the foundation, Hu et al. [18] established a learning and interpretation strategy that revealed a critical biomarker for autism spectrum disorder. In a GAT, the aggregation weights of various nodes are computed in accordance with the importance of neighboring nodes to the central node and introduced into an attention mechanism, helping to better analyze non-Euclidean data [19]. Edge features reflect the correlation between adjacent nodes and contain important information regarding the graph. However, only node features and graph topology are considered in GATs, and edge features are not fully incorporated [20]. For the classification task of SZ, the lack of connectivity information between brain regions can result in suboptimal classification accuracy of the model. Additionally, several studies have demonstrated that a brain region typically interacts with only a few brain regions [21, 22]. Using the full connectivity of brain regions as a brain connectivity network may contain a large number of spurious or weak connections, which could reduce the classification accuracy of the classifier.

To overcome these limitations, this study proposed a GCN architecture based on a sparse interaction mechanism to construct effective sparse connections while utilizing Multi-SIGATnet to update the features, and the generated higher-order features are mapped to the classification target space to achieve the classification of schizophrenia. The significant contributions of this study are as follows.

1)In comparison to the normal multimodal fusion approach, this method considers the interaction between brain regions in addition to fusing traditional neuroimaging features, resulting in the construction of more comprehensive features by using complementary information across modalities.

2)A new sparse GAT model framework was proposed for simultaneously aggregating the local and global features of brain networks. Unlike the original GAT method, the Multi-SIGATnet model injects higher-order edge weights into the node-update process, providing more powerful feature representation.

3)A sparse interaction mechanism was designed to identify significant higher-order interaction features. This method effectively eliminates redundant features and constructs higher-order interaction features by measuringthe cor-relation between brain areas. Consequently, it improves the model performance while reducing computational costs.

## II. METHODS

### A. Data acquisition and preprocessing

The model was evaluated using two freely available datasets, namely, the Center for Biological Research (COBRE, available at http://fcon 1000.projects.nitrc.org/indi/retro/cobre.html) and the University of California Los Ange-les (UCLA, available at https://openneuro.org/datasets/ds000030/versions/1.0.0) datasets.

### 1) Dataset acquisition

The COBRE dataset consists of 147 subjects, including 72 SZ and 75 healthy subjects. Each subject obtained informed consent according to the regulations of the Office of Human Research Protection at the University of New Mexico, details of which can be found at http://cobre.mrn.org/. In this work, 6 subjects were manually excluded from the acquired raw data, as they were not suitable for the data inclusion criteria of this study due to unclear labels or abnormal scan parameters. Therefore, raw anatomical and resting-state functional MRI data and other clinical information from 72 patients with SZ and 69 healthy controls were used from the COBRE dataset. A 3T Siemens Trio scanner was used to obtain whole-brain images of all individuals. sMRI was acquired using a multiple echo (MPRAGE) sequence, the scanning parameters of which were as follows: echo time (TE) = [1.64, 3.5, 5.36, 7.22, 9.08], repetition time (TR) = 2530 ms, inversion time (TI) = 900 ms, flip angle =7°, field of view (FOV) = 256∗256 mm, slab thickness = 176 mm, voxel size = 1∗1∗1 mm. Resting-state functional MRI data were collected using full-space echo-planar imaging in a single shot, the scanning parameters of which were as follows: TR = 2000 ms, TE = 29 ms, voxel size= 334 mm, matrix size = 64∗64, FOV = 192mm∗192 mm, and flip angle = 75°. Detailed parameter information can be found in the COBRE database.

The UCLA dataset provides anatomical sMRI and fMRI data for each subject, including 138 healthy subjects and 58 schizophrenic subjects. The dataset excludes subjects who have neurological disorders, a loss of consciousness, used psychotropic drugs, and a medical history of major psychiatric disorders. Through screening, we used a dataset of 139 subjects, including 89 healthy subjects and 50 patients with SZ. In accordance with the protocols authorized by the UCLA Institutional Review Board, each participant provided written informed consent. All datasets were acquired using a 3T Siemens Trio scanner. The precise scanning parameters for the sMRI dataset were as follows: TR = 1.9 s, TE = 2.26 ms, slice thickness = 1 mm, matrix size = 256∗256, and FOV = 250mm∗250 mm. The scan parameters of the functional MRI dataset were as follows: TR = 2 s, TE = 30 ms, flip angle = 90°, matrix size = 64∗64, slice thickness = 4 mm, and FOV = 192∗192 mm. Detailed parameter information can be found in the UCLA database. More details of the datasets are shown in Table 1.

Table 1
Demographic and clinical information of the dataset

| Dataset | Lable | Gender(n,M/F) | Age(Mean.SD) |
|---|---|---|---|
| COBRE | HC | 72(49/23) | 35.791 ± 11.657 |
|  | SZ | 69(57/12) | 38.0 ± 13.559 |
| UCLA | HC | 89(46/43) | 30.393 ± 8.253 |
|  | SZ | 50(38/12) | 36.46 ± 8.787 |

*2) Dataset preprocessing*

Resting-state fMRI data were preprocessed using the MATLAB toolkit [23]. Specifically, the preprocessing included the following steps: 1)The first ten time points were removed to ensure that the fMRI data achieved a steady state[17]. 2)Time correction was conducted. 3)Head movement correction was conducted to alleviate the influence of head movement in the scanning process. 4)Data from individual spaces of different subjects were registered to a standard space to address differences in brain morphology and inconsistencies in scan spatiotemporal locations among different subjects. 5)A Gaussian smoothing kernel was used for spatial smoothing. After preprocessing, based on the brain template of Destrieux [24], the fMRI volume was divided into 148 ROIs, and the important biomarker values of each ROI were extracted according to previous studies [25, 26], namely, the amplitude of low-frequency fluctuations (ALFF), functional connection strength (FCD), four-dimensional (spatiotemporal) consistency of local neural activities (FOCA), regional homogeneity (ReHo), and fractional amplitude of low-frequency fluctuation (fALFF).

All preprocessing and feature extraction steps were performed using FreeSurfer [27], preprocessing including skull removal, bias field correction, and segmentation of the brain. After preprocessing the MRI data, the following five features were calculated for each ROI from the segmented cortex [28], namely, the surface area (SurfArea), gray matter volume (GrayVol), average thickness (ThickAvg), thickness standard (ThickStd), and integrated rectified mean curvature (MeanCurv).

*B. The proposed method*

A novel multimodal sparse graph convolution architecture for the auxiliary diagnosis of SZ was proposed in this study, the overall framework of which is shown in **Fig. 1**. Because the size and shape of each subject's brain are different, data preprocessing was performed on the MRI data to eliminate differences between the data. The detailed process is described in Section ⅠⅠ.A.2. After preprocessing the MRI data, we extracted important features of each brain region in the fMRI and sMRI data and constructed a multi-modal brain topology with important neuroimaging features of brain regions as node features and correlation coefficients of brain regions as edge features. This brain topology was used for downstream classification tasks.

As a single brain region usually communicates with only a few other brain regions [21]. Consequently, the con-structed multi-modal brain topology was sparsed to better simulate the brain network structure. First, an asymmetric convolutional network was used in the edge feature matrix to enhance the feature information of strong connections and weaken the feature information of weak connections, thereby enhancing the distinction between salient and redundant features. A sparse interaction strategy was then used to retain important features and eliminate redundant features, reducing the computational cost while improving the model performance. Finally, the generative sparse higher-order brain network was passed into the main model. The multi-head attention strategy could calculate the aggregate weight of a neighboring node based on its importance to the central node, which was helpful for the analysis of SZ. Consequently, a multi-head GAT was adopted to simultaneously update and aggregate the local and global features of the brain network, and the features of the model output were fed into a classifier for disease prediction.

*1) Graph construction*

MRI-based disease classification tasks can be classified into two categories based on how a graph is constructed, namely, node-level and graph-level classification. Each subject is typically viewed as a node in the node-level classification task, the biological features of each subject being vectorized and set as the node information of the brain network. The similarity measurement method can then be used to calculate the correlation between each node as the adjacency matrix of the brain network [29, 30]. The constructed graph is input into the network model such that nodes with the same label are classified similarly. Although the composition strategy that treats each subject as a node performs well, it still has certain limitations in considering the mutual influence of various brain regions. The graph-level classification task can fully consider the topological structure of the brain and construct the entire brain connections of the subjects as an undirected brain.graph $G = (X, A, E)$, where $X = \{h_1, h_2, ... h_N\}^T$ denotes the feature set of the node, $h_i \in \mathbb{R}^F$ represents the feature vector of the $i$-th node, $A \in \mathbb{R}^{N*N}$ denotes the connection relationship between brain regions, $E = \{e_{ij}\} \in \mathbb{R}^{N*N*D}$ denotes the feature set of the edge, $F$ and $D$ denote the attribute dimensions of the nodes and edges. In this study, a graph-level strategy was used to construct the brain network. To gain multi-modal information, fMRI and sMRI can be fused through network construction to achieve the advantages of complementing information between modes to help predict labels. The detailed neuroimaging information for constructing the multimodal brain network is presented in **Table 2**.

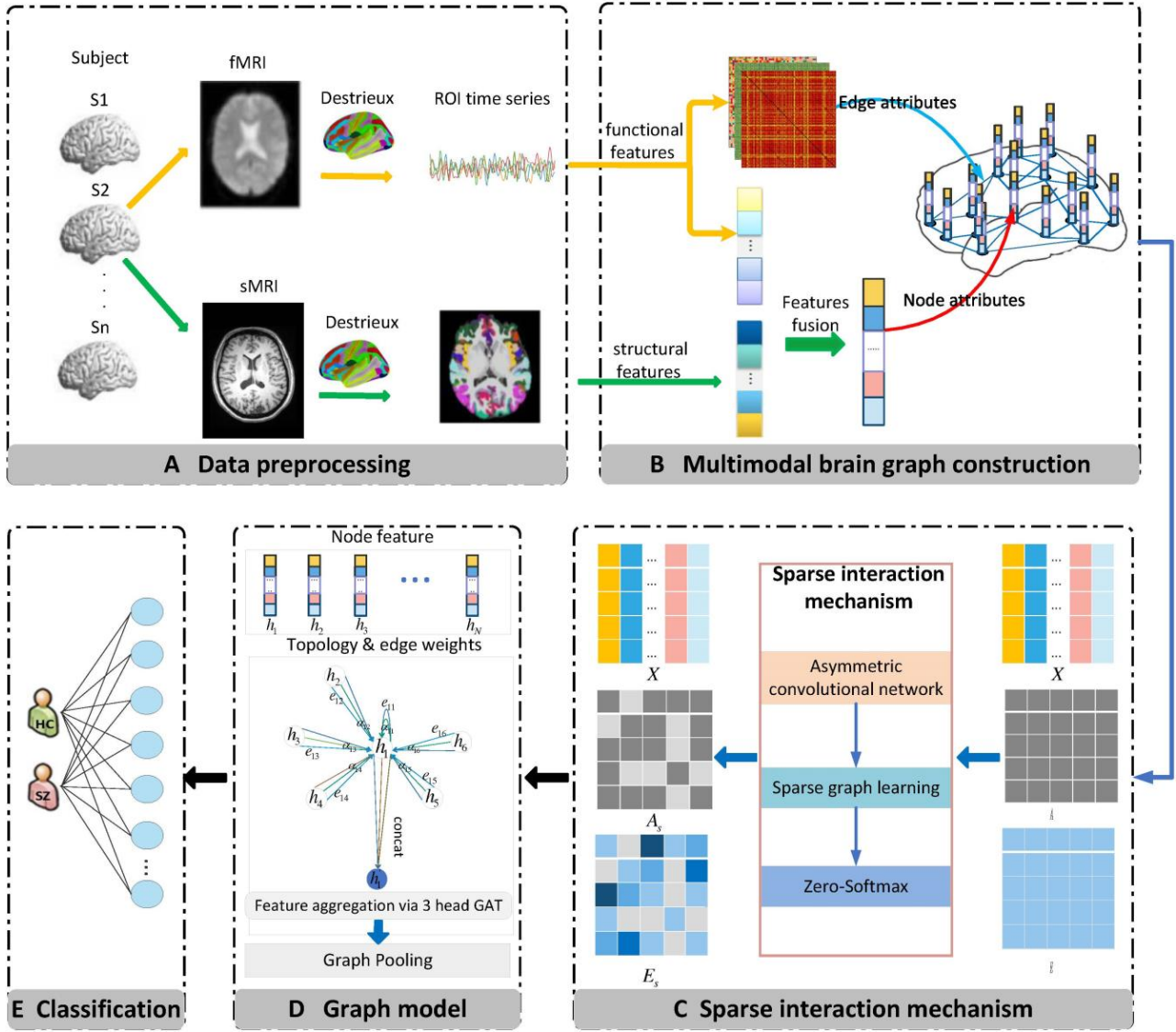

Fig.1. Overall framework flow chart of the Multi-SIGATnet for the classification of schizophrenia. (A) The data preprocessing and feature extraction process. (B) The fMRI and sMRI bioimaging features extracted in stage A are fused by concatenation as node information, the correlation coefficients calculated based on Pearson correlation analysis being used as edge features, and a multimodal brain topology network is constructed by feature embedding. (C) In order to better simulate the structure of brain network, a sparse interaction mechanism is constructed to sparsity the brain network and obtain higher-order interaction features. (D) Based on the multi-head GAT, the local and global features of the brain network are aggregated at the same time, and the central node information is updated. (E) Project the generated higher-order features into the classification target space for disease classification.

Table 2
Biometric information of the graph network.

|  | fMRI | sMRI | SUM |
|---|---|---|---|
| Node features | ALFF | SurfArea | 12 |
|  | fALFF | GrayVol |  |
|  | global_FCD | ThickAvg |  |
|  | local_FCD | ThickStd |  |
|  | longRange_FCD | MeanCurv |  |
|  | FOCA | - |  |
|  | ReHo | - |  |
| Edge features | Pearson correlation coefficient | - | 3 |
|  | Spearman correlation coefficient | - |  |
|  | Minkowski similarity coefficient | - |  |

**Node Features:** The brain is split into 148 ROIs based on the Destrieux template, and each ROI is defined as a graph node. Seven fMRI bioimaging features and five sMRI features are extracted from each ROI, and the information from the two modalities is fused by concatenation as node features. Therefore, for each node, its feature is a vector of $12*1$, and the size of the node feature matrix constructed for each subject is expressed as $148*12$.

**Edge features:** The fMRI brain image is composed of multiple volumes and is a 4D time series. All the signal values of a brain region (such as an ROI) in the sequence are extracted and arranged in the order of scanning time to form a time sequence that changes over time. In this study, the average time series of each ROI in the fMRI data was extracted after preprocessing, and the correlation between the ROIs was calculated based on Pearson and Spearman correlation analyses[31, 32], and the similarity between the two ROIs was calculated using the Minkowski distance[33].

Pearson correlation calculation method is as follows：

$$rij = \frac{\sum_{k=1}^{K}(S_{ik}-\overline{S}_{ik})(Sjk-\overline{S}jk)}{\sqrt{\sum_{k=1}^{K}(S_{ik}-\overline{S}_{ik})^2}\sqrt{\sum_{k=1}^{K}(S_{jk}-\overline{S}_{jk})^2}} \quad (1)$$

where $r_{ij}$ denotes the Pearson correlation coefficient for time series $S_i$ and $S_j$, $K$ is the number of time points in the time series, $S_{ik}$ is the blood oxygen level dependent signal value at the $k$ th time point, and $\overline{S}_{ik}$ is the average signal value at $K$ time points on $S_i$. Spearman correlation is calculated as shown follows:

$$\rho_{ij} = 1 - \frac{6\sum_{k=1}^{K}d_k^2}{K(K^2-1)} \quad (2)$$

where $\rho_{ij}$ represents the Spearman correlation coefficient of time series $S_i$ and $S_j$, $d_k$ and is the rank difference between $S_{ik}$ and $S_{jk}$. Minkowski similarity is calculated as follows:

$$mij = \sqrt[p]{\sum_{k=1}^{K}|S_{ik}-S_{jk}|^p} \quad (3)$$

where $mij$ is the similarity coefficient of time series $S_i$ and $S_j$, and $p$ is a constant ($p=2$). Finally, the three metrics are fused as edge features by element-wise summation. Therefore, the final edge feature $e_{ij}$ can be expressed as follows:

$$e_{ij} = add(r_{ij}, \rho_{ij}, m_{ij}) \quad (4)$$

Here, add indicates that the sum of the corresponding position elements is used for feature information fusion[34].

*2) Sparse Graph Attention Network*

Graph message passing: The GNN aims to update the feature information of the central node by aggregating features from neighboring nodes [35]. GCN methods can be divided into two categories, that is, spectral-domain and spatial-domain based methods. The spatial-domain based approach represents graph convolution as the aggregation of feature information from neighbors as shown in **Fig.2**. From a graph signal processing perspective, spectral-based methods define the graph convolution in the spectrum domain. Whether spatial GCNs or spectral GNNs, a common convolutional layer typically comprises two stages, that is, message passing $M_t$ and feature updating $U_t$ [36]. The convolution process of the $l$-th layer be expressed as follows:

$$m_i^{l+1} = \sum_{j\in\mathcal{N}(i)} Mt(h_i^l, h_j^l, e_{ij}) \quad (5)$$

$$h_i^{l+1} = U_t(h_i^l, m_t^{l+1}) \quad (6)$$

where $h_i^l$ denotes the feature vector of node $i$ at layer $l$, $e_{ij}$ denotes the connection weight of node $i$ and $j$ node, hid-den layer $h^l$ updates each node in the graph based on the message $m_i^{l+1}$, and $\mathcal{N}(i)$ denotes the neighboring node of the node $i$. In contrast to the GCN, the GAT applies an attention mechanism to a spatial-domain–based GNN and updates node information by focusing on the representation of neighboring nodes, which is helpful for SZ classification tasks. However, the GAT updates node features based only on the topology of the brain network, ignoring the importance of edge features in the message passing process. The lack of connectivity information can affect the performance of the classification model. Consequently, in this study we proposed a new graph attention method for SZ identification tasks.

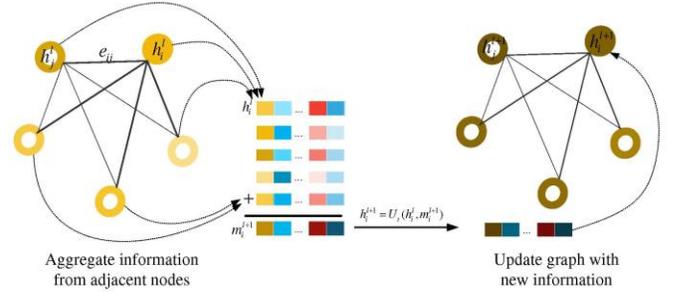

**Fig.2**. Node message passing flow chart. Message passing is a process of gathering data from nearby nodes to update the information of the central node. It includes three steps: (A) Transform the neighboring node information based on the edge weight; (B) Aggregate the information of neighboring nodes to the central node. The aggregation function can be sum, max, mean, and other aggregation methods. The update method of the sum aggregation function is shown here;(C) Update the node information of the $l+1$ layer based on the own node features and neighboring node features of the th $l$ layer.

**Graph sparse representation**

Typically, an individual brain region communicates with only a small number of other brain regions. When certain sparse constraints are imposed on the connectivity network, redundant connections can be eliminated from a set of noisy connections. In this study, the brain network was separated based on the structural characteristics of the brain as showm in **Fig.3**. $E$ depicts the intensity of the connectivity between brain regions. To make it easier for downstream tasks to distinguish salient features and redundant features, asymmetric convolutional networks [37] are used for edge features to enhance salient features and weaken redundant features to obtain higher-order interactive features, which is described as follows:

$$\begin{aligned}H_{row}^{(l)} &= Conv(E^{(l-1)}, K_{1\times k}^{row})\\H_{col}^{(l)} &= Conv(E^{(l-1)}, K_{1\times k}^{col})\\H_{all}^{(l)} &= Conv(E^{(l-1)}, K_{1\times k}^{all})\\H^{(L)} &= \delta(H_{row}^{(l)} + H_{col}^{(l)} + H_{all}^{(l)})\end{aligned} \quad (7)$$

where $H_{row}^{(l)}$ and $H_{col}^{(l)}$ denote the row-based and column-based asymmetric convolution feature maps at the l-th layer, respectively. denotes a nonlinear activation function, and $K_{1\times k}^{row}$, $K_{1\times k}^{col}$ and $K_{1\times k}^{all}$ denote convolution kernels of size $1\times k$, $k\times 1$ and $k\times k$ respectively. To obtain the same sized feature inputs and outputs, zero-padding operations are performed on all convolutions. The output of the last layer of the asymmetric convolution is a high-order interactive feature H. To reduce the impact of weak brain connections on the classification and diagnosis performance of SZ, after obtaining the high-order interaction feature matrix, the sparse learning strategies are set to realize the sparseness of the edge feature matrix. The hyperparameter is set as $\xi \in \{0, 0.3, 0.5, 0.7, 1\}$ to perform element-wise thresholding on $H$. That is, $H(i,j) \geq \xi$, the element of $(i,j)$ is set to $H(i,j)$, otherwise it is 0.

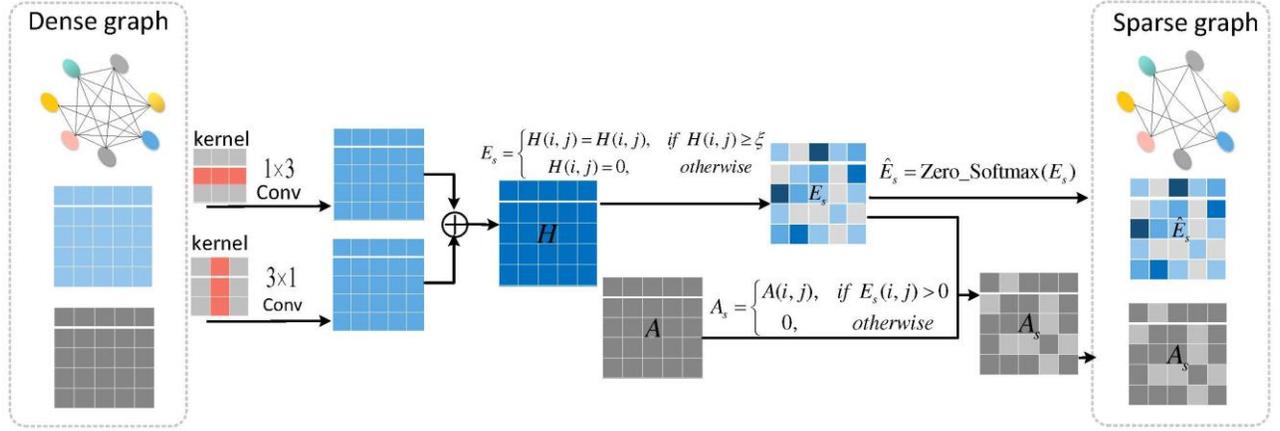

Fig.3. A flow chart of the sparse interaction mechanism. The rows and columns of the edge feature matrix are first combined based on asymmetric convolution to boost the strength of prominent features while highlighting the backbone information to obtain high-order interactive features. The mask matrix is then set to eliminate redundant connections based on the sparse learning strategy, to generate a high-order sparse graph.

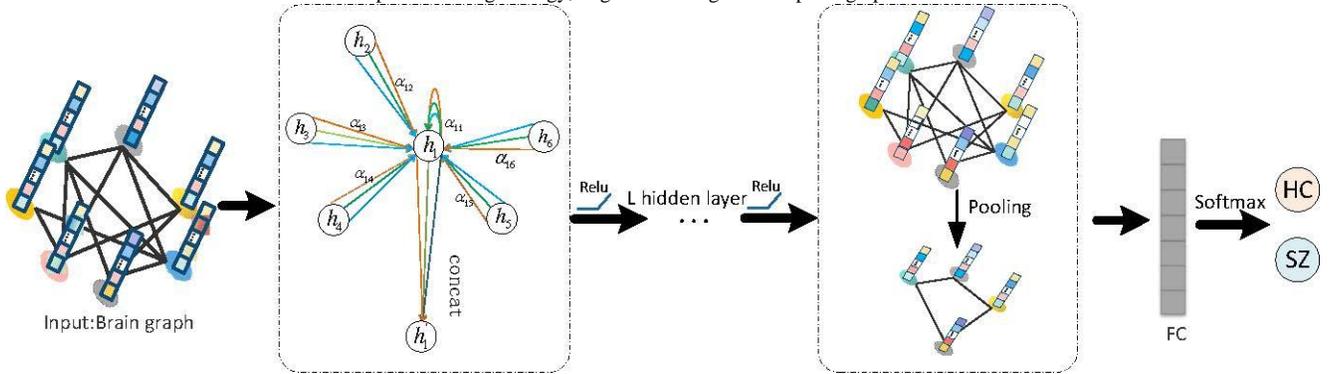

Fig.4. The architecture of the graph convolution. The framework consists of three main parts. The convolution stage contains three convolution layers, each of which adopts a multi-head graph attention network to simultaneously aggregate the local and global features of the brain network to update node information. Each convolutional layer is followed by a Relu activation function. To alleviate the overfitting problem, the global average pooling method is used for downsampling in the pooling stage. Lastly, a full connection layer with Softmax is used to classify schizophrenia.

$$Es = \begin{cases} H(i,j) = H(i,j), & \text{if } H(i,j) \geq \xi \\ H(i,j) = 0, & \text{otherwise} \end{cases} \quad (8)$$

Previous studies have shown that the normalization of the feature matrix is vital for the normal operation of the graph neural network[38]. Therefore, after obtaining the high-order sparse edge feature matrix, Zero_Softmax[39] is used to perform row normalization on the feature matrix as opposed to directly using Softmax to normalize the adjacency matrix. Softmax may cause the input zero value to become a non-zero value output, making the generated sparse matrix return to a dense matrix again. In this case, non-interacting brain regions are forced to interact again. In order to maintain the sparsity of the feature matrix, this study uses Zero_Softmax to perform row normalization on the high-order sparse edge feature matrix, compressing the value of the feature matrix to [0; 1], to eliminate the influence of singular data on the convergence of the model. Let us consider a vector $e = [e_{i1}, e_{i2}, ..., e_{i\mathcal{D}}]$ of a certain dimension of the feature matrix is given, then

$$E_s = Zero\_Soft\max(E_s) = \frac{(\exp(e_{ij})-1)^2}{\sum_k^{\mathcal{D}}(\exp(e_{ik})-1)^2 + \varepsilon} \quad (9)$$

where $\mathcal{D}$ represents the dimension of the input feature and $\varepsilon$ is a small constant used to ensure numerical stability and avoid division by zero. The normalized sparse edge feature matrix $E_s$ can be expressed as follows:

$$E_s = Zero\_Soft\max(E_s) \quad (10)$$

A is a fully connected adjacency matrix that indicates that all brain regions interact. Based on the importance of the edge features of the brain network, the corresponding sparse adjacency matrix As can be constructed, which is set to zero for redundant connectivity and one for the rest. Consequently, the obtained sparse undirected graph can be expressed as follows.

$$As = \begin{cases} A(i,j), & \text{if } E_s(i,j) > 0 \\ 0, & \text{otherwise} \end{cases} \quad (11)$$

**Edge Graph Attention Convolution**

**Node representation learning:** The GAT can learn sparse graph information and can be used for the research of classification tasks. To improve the performance of the algorithm, inspired by [20] and [28], the edge weight $e_{ij}$ is introduced into GAT. The input of Multi-SIGATnet is a set of node features, $X = \{h_1, h_2, ..., h_N\}, h_i \in \mathbb{R}^F$. The graph attention layer uses the self-attention mechanism to gather the neighborhood node information related to the target node to update the center node. The graph convolution process is shown in **Fig.4**. The calculation method of the attention coefficient $\alpha_{ij}$ is as follows:

$$\alpha_{ij} = \frac{\exp(\text{Leaky}\text{Re}lu(a^T[Wh_i \| Wh_j])e_{ij})}{\sum_{k \in \mathcal{N}(i)} \exp(\text{Leaky}\text{Re}lu(a^T[Wh_i \| Wh_k])e_{ik})} \quad (12)$$

where $a \in \mathbb{R}^{2F'}$ is a weight vector, $*^T$ represents the transpose, and $W \in \mathbb{R}^{F' \times F}$ is a learnable linear transformation initialized with glorot_uniform, which maps the feature vector of each node from dimension $F$ to dimension $F'$, improving the expressive ability of features. $\|$ is used to concatenate the mapped feature vectors. The value after linear change is activated by the LeakyRelu activation function, and the negative slope is set to 0.2. To obtain a more reliable attention learning process, the multi-head attention mechanism learning approach was used in this study to update the node information. The feature representation method of node i based on the multi-head attention mechanism can be expressed as follows:

$$h_i' = \|_{t=1}^T \sigma(\sum_{j \in \mathcal{N}(i)} \alpha_{ij}^t W^t h_j) \quad (13)$$

The results calculated by $T$ independent attention mechanisms are concatenated as the features of the updated nodes. $T$ denotes the number of attention mechanisms, $\|$ represents concatenation, and $\sigma(\cdot)$ denotes the activation function, where $h_i'$ is the updated feature vector. During information aggregation, in addition to considering the attention coefficients of neighboring nodes, the model also adds a self-attention coefficient, which can be calculated as follows.

$$h_i'' = \|_{t=1}^T \sigma(\alpha_{ii}^t W^t h_i + \sum_{j \in \mathcal{N}(i)} \alpha_{ij}^t W^t h_j) \quad (14)$$

where $\alpha_{ii}^t$ is the self-attention coefficient and $\|$ represents concatenation.

**Pooling layer:** Pooling is an important downsampling operation in traditional CNNs. For graph-level classification, the number of nodes and feature dimension of each node are generally relatively large, and the data dimension can be reduced by pooling, thereby alleviating the problem of overfitting. In a GCN, the convolutional layer does not change the structure of the graph but only updates the information at the nodes. Therefore, it is necessary to set up a pooling layer after the convolutional layer. Graph-pooling methods can be grouped into two categories—that is, the global pooling method, which considers global features and can be used to handle graphs of different structures [40, 41], and the hierarchical pooling method, which learns to capture hierarchical information that is crucial to the graph structure, such as TopKPooling [42] and SAGPooling [43]. In this study, the global pooling approach was used, the graph-level output being returned by averaging the node features in the node dimension, which can be expressed as follows:

$$z^k = \text{mean}\{h_i^k : i = 1, ..., N^k\} \quad (15)$$

where $h_i^k$ represents the feature vector of the *i*-th node at layer *k*, after which *z* can then be sent to the fully connected layer for the final classification.

3) *Loss function*

The loss function is used to measure the error between the predicted value of the sample and the real value. BinaryFocalCrossentropy [44] can solve the problem of sample imbalance. In this study, the BinaryFocalCrossentropy loss function is used for the schizophrenia classification training task, and its definition is shown in Eq. (16):

$$Loss = -\frac{1}{N}\sum_{i=1}^N (\alpha y_i(1-p(y_i))^\gamma \log(p(y_i)) + (1-\alpha)(1-y_i)p(y_i)^\gamma \log(1-p(y_i))) \quad (16)$$

where $N$ denotes the number of samples, $y_i$ denotes the label of the *i*-th sample, if the sample is a positive example, the value is 1, otherwise the value is 0. $p(y_i)$ denotes the probability that the *i*-th sample is predicted to be positive. $\alpha \in [0,1]$ is a weighting factor used to solve the problem of unbalanced sample size. $\gamma \geq 0$ is an adjustable focusing parameter, which can reduce the loss contribution of easy-to-distinguish samples, thereby increasing the loss ratio of difficult-to-distinguish samples.

III. EXPERIMENTS

*A. Implementation details and model evaluation*

The Multi-SIGATnet algorithm is built based on the tensorflow framework. The experiment uses a 5-fold cross-validation strategy to test the stability of the model. To ensure fairness, the following parameters are used for all experiments: learning rate = 0.001, batch_size = 32, epochs=150. The optimizer is adam. Four evaluation metrics commonly used in classification tasks are used as the evaluation indicators of the Multi-SIGATnet model results: classification accuracy (ACC), sensitivity (SEN) and specificity (SPE), and F1 score[45]. Experiments use NVIDIA GeForce RTX 2080 devices for training.

*B. Comparison with traditional methods*

The proposed method was compared with other existing methods to verify the validity of the Multi-SIGATnet algorithm in SZ classification, including (1) machine learning methods: random forest (RF), support vector machine (SVM), (2) convolutional neural network (CNN), and (3) GAT. All algorithms were trained based on the fMRI and sMRI fusion data. The CNN comprises three convolutional layers, each of which can be activated by the Relu function, where the size of the convolution kernel is $3*3$. The GAT adopted the same parameters as the proposed method.

**Table 3** shows the classification results on the COBRE and UCLA datasets. The experimental results show that on the two datasets, the proposed method is significantly superior to other methods in the classification performance of four performance metrics. For the COBRE dataset, The accuracy, sensitivity, specificity and F1 score of the method proposed in this study achieve 81.9%, 86.8%, 80.5%, and 80.8%, respectively. Compared to the traditional method, the proposed model achieves the highest sensitivity, which shows that the method can effectively improve the dis-crimination between SZ and HC. When verified on the UCLA dataset, compared with the GAT method, the accuracy, sensitivity, specificity and F1 score of the proposed method increased by 5.5%, 6.4%, 2.0%, and 8.7%, respectively, leading to better classification performance. In the overall analysis, compared to the existing methods, the proposed method has certain advantages in identifying SZ.

Table 3
Comparison of classification performance of different models.

|  |  | ACC(%) | SEN(%) | SPE(%) | F1(%) |
|---|---|---|---|---|---|
| COBRE | SVM | 55.3 | 45.5 | 65.6 | 49.1 |
|  | RF | 63.1 | 55.6 | 71.6 | 59.1 |
|  | CNN | 74.4 | 72.9 | 76.1 | 73.6 |
|  | GAT | 77.3 | 81.5 | 73.2 | 76.4 |
|  | Multi-SIGATnet | 81.9 | 86.8 | 80.5 | 80.8 |
| UCLA | SVM | 57.4 | 36.3 | 70.4 | 55.2 |
|  | RF | 61.0 | 60.8 | 63.0 | 60.4 |
|  | CNN | 67.6 | 41.9 | 87.4 | 51.7 |
|  | GAT | 70.3 | 55.7 | 80.9 | 58.6 |
|  | Multi-SIGATnet | 75.8 | 62.1 | 82.9 | 67.3 |

### *C. Ablation studies*

In this study, a series of ablation experiments was designed to demonstrate the validity of each Multi-SIGATnet model module. From the experimental findings, it can be concluded that each component contributed to improving the classification capacity of the proposed methodology.

#### *1) Effectiveness of edge features*

The edge features of the brain network reflect the intensity of communication between various brain areas, and the influence of edge features on the classification performance is considered in Multi-SIGATnet. As can be seen from **Table 4**, in the COBRE data set and UCLA data set, the embedding of edge features significantly improved the classification accuracy of the model, which verifies the effectiveness of edge features for schizophrenia classification. Through the experimental analysis, it can be found that when the single Pearson correlation coefficient is used as the edge feature, the accuracy of the two datasets is improved by 1.2% and 0.8% respectively. When the fused data of the three metrics were used as the edge features, the classification performance of the model was the best, with the accuracy of 79.4% and 72.1%, respectively. In Table 4, P represents Pearson correlation, S represents Spearman correlation, and M represents Minkowski similarity.

#### *2) Effectiveness of Multi-Head Learning*

During the network model training procedure, the selection of the number of attention heads is an important hyperparameter. In this experiment, the representation ability of multi-head learning was studied from the perspective of classification accuracy. **Table 5** summarizes of the findings.

Table 4
Effectiveness verification of edge features.

|  |  | ACC(%) | SEN(%) | SPE(%) | F1(%) |
|---|---|---|---|---|---|
| COBRE | GAT | 77.3 ± 4.8 | 81.5 ± 3.4 | 73.2 ± 6.2 | 76.4 ± 5.3 |
|  | GAT-P | 78.5 ± 3.4 | 82.3 ± 3.2 | 73.8 ± 4.5 | 77.8 ± 2.4 |
|  | GAT-P&S | 79.1 ± 2.6 | 83.4 ± 3.1 | 74.5 ± 5.2 | 78.2 ± 3.1 |
|  | GAT-P&S&M | 79.4 ± 3.6 | 83.8 ± 5.1 | 75.8 ± 6.1 | 78.8 ± 2.3 |
| UCLA | GAT | 70.3 ± 4.3 | 55.7 ± 8.9 | 78.9 ± 7.1 | 56.5 ± 7.3 |
|  | GAT-P | 71.1 ± 4.5 | 56.2 ± 6.5 | 79.2 ± 5.6 | 56.9 ± 7.1 |
|  | GAT-P&S | 71.9 ± 3.8 | 57.6 ± 7.2 | 79.3 ± 4.8 | 57.2 ± 6.9 |
|  | GAT-P&S&M | 72.1 ± 6.5 | 58.0 ± 7.6 | 79.6 ± 8.2 | 57.8 ± 8.96 |

It can be observed that the network performance improves with the increase of the amount of attention heads. For the COBRE and UCLA datasets, when there are three attention heads, the model achieves the best classification performance, 80.5% and 74.6%, respectively; Compared with T=1, the accuracy is improved by 1.1% and 2.5%, respectively. This shows that using multi-head attention can learn features from multiple scales and enhance feature information. From the experimental results, it can be found that when the number of heads exceeds three, the accuracy of both COBRE and UCLA datasets decreases slightly. The analysis shows that in the process of feature iterative update, the features updated based on different attentions are fused by concatenation. Therefore, when there are too many heads, a large amount of redundant information is generated, which affects the classification performance.

Table 5
Effectiveness verification of multi-head learning.

|  |  | ACC(%) | SEN(%) | SPE(%) | F1(%) |
|---|---|---|---|---|---|
| COBRE | T=1 | 79.4 | 82.8 | 72.8 | 71.6 |
|  | T=2 | 79.8 | 78.1 | 80.5 | 78.6 |
|  | T=3 | 80.5 | 78.5 | 82.3 | 79.8 |
|  | T=4 | 77.1 | 79.3 | 75.0 | 76.9 |
|  | T=5 | 73.7 | 72.8 | 76.7 | 72.4 |
| UCLA | T=1 | 72.1 | 47.5 | 82.8 | 53.4 |
|  | T=2 | 73.9 | 61.1 | 78.3 | 61.8 |
|  | T=3 | 74.6 | 64.9 | 78.5 | 62.3 |
|  | T=4 | 72.2 | 57.3 | 76.5 | 59.2 |
|  | T=5 | 69.4 | 58.5 | 76.2 | 57.7 |

#### *3) Effectiveness of sparse learning*

**Table 6** shows the effectiveness of the brain topology with sparse interactions. 1) Multi-SIGATnet-1: hyperparameter=1, which implies that each brain interval is independent of another, having no mutual influence; 2) Multi-SIGATnet-2: =0.7, which implies an extremely sparse interaction in the brain regions; 3) Multi-SIGATnet-3: the hyperparameter =0.5; 4) Multi-SIGATnet-4: =0.3, which implies a relatively dense interaction between the brain regions; 5) Multi-SIGATnet-5: =0, which implies that the brain regions are fully connected. The experimental findings demonstrate that the overall performance of the proposed method reaches a peak when = 0.5. Compared to the fully connected Multi-SIGATnet-5, it achieves 1.4% and 0.9% accuracy improvements on COBRE and UCLA datasets, respectively. Multi-SIGATnet-1 has the lowest performance; its accuracy, sensitivity, and specificity are 5.5%, 9.6% and 5.7% lower on COBRE dataset than Multi-SIGATnet-3, and 5.5%, 6.1% and 3.7% lower on UCLA dataset, respectively, which means that it is necessary to model interactions between brain regions, and properly sparse interactions can indeed lead to improved performance.

Table 6
Effectiveness verification of sparse learning

| Datasets | | Multi-SIGATnet | | | | |
| --- | --- | --- | --- | --- | --- | --- |
| | | Multi-SIGATnet-1 $\xi = 1$ | Multi-SIGATnet-2 $\xi = 0.7$ | Multi-SIGATnet-3 $\xi = 0.5$ | Multi-SIGATnet-4 $\xi = 0.3$ | Multi-SIGATnet-5 $\xi = 0$ |
| COBRE | ACC(%) | 71.2±7.4 | 71.6±2.5 | 75.2±6.0 | 74.4±8.9 | 73.4±5.4 |
| | SEN(%) | 64.3±13.6 | 67.8±7.0 | 61.3±13.8 | 69.9±15.0 | 70.6±12.8 |
| | SPE(%) | 78.5±7.2 | 74.4±8.4 | 79.2±5.8 | 78.3±9.5 | 77.3±11.8 |
| UCLA | ACC(%) | 70.4±6.2 | 71.1±8.5 | 73.4±6.6 | 74.7±8.2 | 72.8±3.4 |
| | SEN(%) | 46.8±11.4 | 58.7±14.3 | 62.7±12.8 | 63.6±10.5 | 57.7±14.5 |
| | SPE(%) | 84.5±6.5 | 78.6±6.9 | 79.0±6.1 | 75.2±11.4 | 82.5±7.4 |

*4) Effectiveness of multimodal data fusion*

**Fig.5** demonstrates the importance of integrating multimodal data by comparing with single modality classification performance. **Fig.5(a)** shows the classification precision on the COBER dataset. It can be found that the multimodal data based on fMRI and sMRI fusion has significantly improved the three evaluation metrics of accuracy, sensitivity and specificity, which are 81.9%, 86.8%, 80.5%, which is 7.1% higher than the accuracy of fMRI, and 15.3% higher than that of sMRI. Like the COBRE dataset, UCLA achieved higher accuracy on the multimodal dataset, as shown in **Fig.5(b)**. Compared with fMRI, its average accuracy, sensitivity, and specificity were improved by 5.0%, 8.7%, 1.6%. Compared with sMRI, it increased by 10.4%, 15.4% and 6.3% respectively. The classification performance of multimodal data on both datasets is shown to be better than the corresponding single modality data, which validates the ability of Multi-SIGATnet to extract complementary information from multiple modal data.

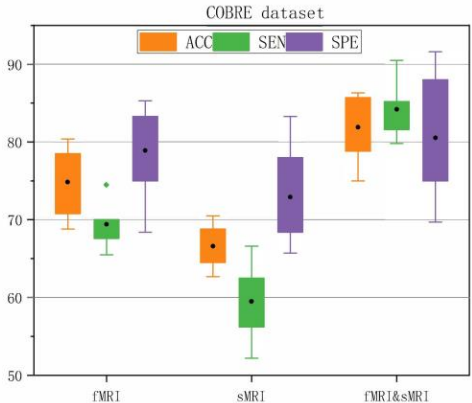

(a)

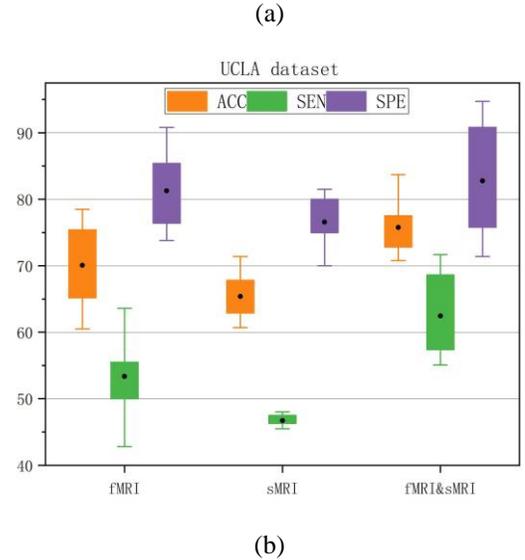

(b)

Figure 5: Comparison of classification performance under different modalities: (a) COBRE dataset, (b) UCLA dataset

### D. Comparison with existing graph neural networks

Based on different pooling methods, the Multi-SIGATnet method was compared with several existing GNNs. **Table 7** and **Table 8** show the experimental outcomes.

ChebConv[46] is a GCN based on spectral methods that implements convolution definitions in the spectral domain based on Laplacian matrixes.

GraphSageConv[47] is a general inductive framework that can generate feature embeddings for previously unseen data by sampling and aggregating features in the local neighborhoods of nodes.

GatedGraphConv[40] is a GNN based on gated recurrent units, which is an extension of the GNN for output sequences.

EdgeConv[48] and ECCConv[49] consider not only the node attributes, but also the edge attributes of the graph. ECCConv uses the dynamic convolution kernel network to generate the edge attributes of the graph network as filter weights and then uses the weights to update the node features. EdgeConv defines the neighboring nodes of the central node based on the k-nearest neighbor method and then extracts the edge features between the center node and the neighboring nodes using a multi-layer perceptron for convolution operations. The classification performance of the Multi-SIGATnet method is better than that of existing GCN algorithms, based on a study of the experimental findings. Specifically, the experiments separately evaluated the classification results of each GCN under five different pooling methods.

Table 7
Comparison of classification performance between the proposed model and existing graph convolutional neural networks on COBRE dataset based on different pooling methods.

| Method | GlobalAttentionPool | | | GlobalAvgPol | | | GlobalSumPool | | | SortPool | | | TopKPool | | |
|---|---|---|---|---|---|---|---|---|---|---|---|---|---|---|---|
| | ACC (%) | SEN (%) | SPE (%) | ACC (%) | SEN (%) | SPE (%) | ACC (%) | SEN (%) | SPE (%) | ACC (%) | SEN (%) | SPE (%) | ACC (%) | SEN (%) | SPE (%) |
| ChebConv | 60.9 | 69.7 | 60.1 | 71.0 | 68.5 | 74.7 | 73.1 | 76.6 | 70.2 | 65.0 | 68.0 | 61.8 | 69.4 | 68.8 | 69.7 |
| ECCConv | 57.4 | 56.7 | 59.7 | 65.6 | 67.0 | 67.1 | 64.8 | 62.5 | 63.9 | 63.1 | 60.2 | 68.7 | 67.4 | 68.3 | 67.1 |
| EdgeConv | 68.0 | 67.1 | 74.2 | 68.5 | 66.4 | 67.7 | 63.9 | 71.4 | 59.5 | 58.6 | 54.4 | 63.1 | 65.3 | 68.6 | 62.2 |
| GateGraph | 52.1 | 51.0 | 51.8 | 55.8 | 47.7 | 55.2 | 54.5 | 61.8 | 50.4 | 56.1 | 48.2 | 67.4 | 51.2 | 49.6 | 59.6 |
| GraphSage | 73.8 | 79.3 | 71.6 | 71.7 | 74.9 | 69.6 | 68.8 | 68.9 | 73.0 | 67.4 | 70.1 | 68.4 | 68.9 | 69.0 | 72.7 |
| GAT | 74.2 | 78.5 | 75.9 | 77.3 | 81.5 | 73.2 | 75.4 | 72.8 | 76.6 | 74.1 | 72.8 | 76.2 | 69.2 | 67.3 | 71.5 |
| Multi-SIGATnet | 78.5 | 78.4 | 76.1 | 81.9 | 86.8 | 80.5 | 78.6 | 76.3 | 77.9 | 74.8 | 73.5 | 76.7 | 72.6 | 69.5 | 73.6 |

Table 8
Comparison of classification performance between the proposed model and existing graph convolutional neural networks on the UCLA dataset based on different pooling methods.

| Method | GlobalAttentionPool | | | GlobalAvgPol | | | GlobalSumPool | | | SortPool | | | TopKPool | | |
|---|---|---|---|---|---|---|---|---|---|---|---|---|---|---|---|
| | ACC (%) | SEN (%) | SPE (%) | ACC (%) | SEN (%) | SPE (%) | ACC (%) | SEN (%) | SPE (%) | ACC (%) | SEN (%) | SPE (%) | ACC (%) | SEN (%) | SPE (%) |
| ChebConv | 70.0 | 52.0 | 83.8 | 69.0 | 32.8 | 89.8 | 62.5 | 42.9 | 72.8 | 68.3 | 38.0 | 84.7 | 64.7 | 28.4 | 76.4 |
| ECCConv | 52.5 | 37.3 | 60.2 | 65.4 | 54.7 | 71.8 | 57.1 | 40.8 | 66.3 | 57.5 | 35.5 | 70.4 | 64.7 | 46.2 | 75.9 |
| EdgeConv | 60.2 | 45.0 | 67.8 | 59.1 | 37.1 | 69.8 | 64.1 | 46.4 | 73.0 | 56.6 | 46.6 | 62.1 | 62.6 | 49.4 | 69.4 |
| GateGraph | 60.5 | 37.4 | 70.1 | 57.8 | 23.7 | 76.3 | 54.5 | 45.3 | 58.6 | 60.8 | 39.8 | 76.7 | 66.1 | 24.9 | 89.9 |
| GraphSage | 72.6 | 72.9 | 81.0 | 65.7 | 39.6 | 80.6 | 68.7 | 46.7 | 79.3 | 66.1 | 47.9 | 77.0 | 66.9 | 16.8 | 93.2 |
| GAT | 70.9 | 57.6 | 66.7 | 70.3 | 55.7 | 80.9 | 71.5 | 52.3 | 80.9 | 68.5 | 50.2 | 77.4 | 67.1 | 52.6 | 71.3 |
| Multi-SIGATnet | 72.5 | 58.2 | 67.4 | 75.8 | 62.1 | 82.9 | 73.2 | 53.8 | 82.1 | 70.2 | 53.5 | 79.2 | 69.6 | 54.3 | 75.2 |

**Table 7** shows the classification performance of the COBRE dataset (the bold font indicating the best classification performance). Furthermore, to analyze the capability of each algorithm, **Fig.6** shows boxplots of the accuracy of the different models under different pooling methods. From the experimental results, it can be found that when GlobalAvgPool is applied as the pooling method, the proposed algorithm achieves the best classification results, with average accuracy, sensitivity and specificity of 81.9%, 86.8% and 80.5%, respectively. Compared to GAT, which is the next best result, the evaluation indicators are improved by 4.6%, 5.3% and 7.3%, respectively. When GlobalSumPool is used as the pooling method, the sensitivity of the Multi-SIGATnet algorithm is slightly inferior to that of the ChebConv method, which is 76.3%; however, the accuracy and specificity are increased by 5.5% and 7.7%, respectively. In the overall analysis, compared to other graph convolutional neural networks, the average accuracy of the proposed model is above 72%. This shows that the Multi-SIGATnet method can effectively enhance the discrimination between SZ and HC, which proves that the model has a stronger feature learning ability.

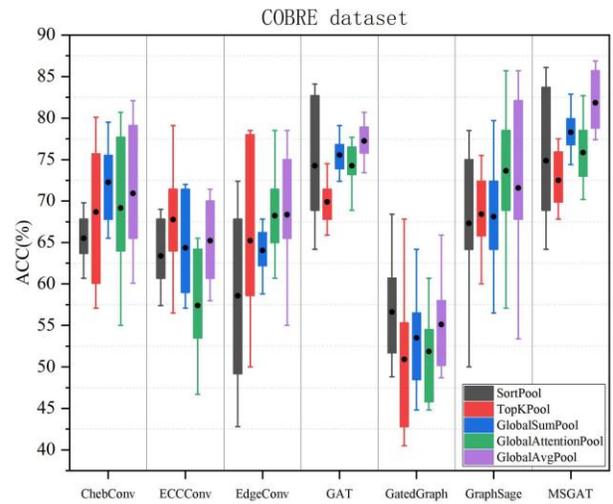

Fig 6: Boxplot of classification accuracy of each model on COBRE dataset

**Table 8** shows the classification performance comparison between the existing GCN and the Multi-SIGATnet algorithm on the UCLA dataset. Clearly, the classification precision on the UCLA dataset is generally inferior to that on the COBRE dataset, owing predominantly to the imbalance of the UCLA data. However, it can be seen from the results that the sensitivity of the Multi-SIGATnet model are superior to those of other convolutional network models, indicating that the Multi-SIGATnet algorithm has a certain anti-interference resistance to the influence of data imbalances. Through experimental analysis, it can be seen that, except for the GlobalAttentionPool method, the average accuracy of the Multi-SIGATnet method is 0.1% lower than that of the GraphSageConv method. Under the other pooling methods, the average accuracy of the proposed model slightly improves. Among them, when GlobalAvgPool is the pooling method, an optimal classification accuracy of 75.8% is achieved, which is 5.5%í18% higher than that of other GCN methods. It can be seen from **Fig.7**, that under the GlobalAvgPool pooling method, the Multi-SIGATnet method achieves the highest global accuracy of 85.7%. The average accuracy of the model under different pooling methods is relatively concentrated, mainly distributed between 69% and 76%, which verifies the stability of the model. Overall, it is clear that the proposed method improves the classification of SZ brain imaging data.

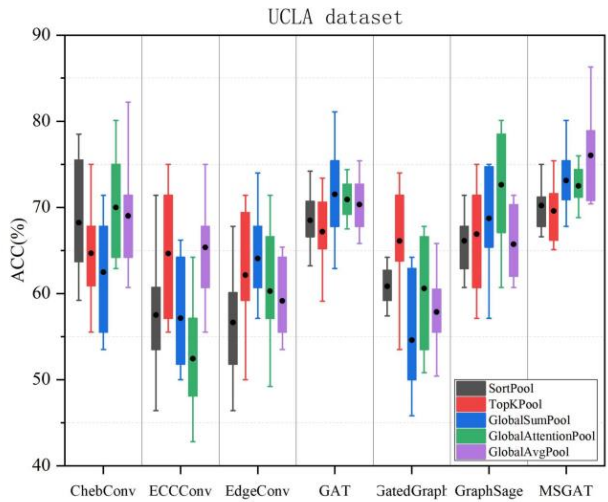

Fig 7: Boxplot of classification accuracy of each model on UCLA dataset

### IV.  Discussion

A GNN can directly capture the topological structure of the brain network compared to the traditional CNN model, making it preferable for mining the local and global information of the brain as it shows tremendous promise in the analysis of brain networks [50]. GNN convolution operators can be created in either the spatial or spectral domain, but because convolution in the spectral domain requires complex feature decomposition calculations, the calculation cost can be relatively high. Therefore, an increasing number of GNNs have designed convolution operators based on spatial methods. The Multi-SIGATnet model inherits the spatial domain's convolutional advantages, which increases the classification effectiveness of the model while lowering its computational cost.

The Multi-SIGATnet model fuses data from two modalities, namely, fMRI and sMRI. Compared with the single modal method, multimodal data on the COBRE dataset achieves an accuracy improvement of 7.1% (fMRI) and 15.3% (sMRI), respectively, the UCLA dataset improving by 5.0% (fMRI) and 10.4% (sMRI), respectively. This proves that combining multimodal data can yield a more comprehensive feature representation for SZ classification. Unlike the traditional GAT algorithm, this study embeds the connection weights of brain intervals when constructing a multimodal brain network, and constrains feature updates according to the topological features and connection weights of the high-order network. As can be seen from Table 4, using the brain interval connection weight as the edge feature information to embed the multimodal data, the classification performance of the model has been improved, which verifies that interaction features are of great importance for the correct classification of SZ.

Based on the sparse interaction mechanism, this study tested the effect of brain network sparsity on classification performance. From the results shown in Table 6, it can be concluded that the classification performance of the model is improved when certain sparsity restrictions are imposed on the brain network, which proves that the sparse inter-action mechanism can filter out weak or false connections in the brain network. However, when there is excessive sparseness (for example, there is no interaction between various brain regions) accuracy decreases, indicating that for the classification of SZ, modeling the interaction relationship based on brain regions is necessary.

Although the Multi-SIGATnet method shows promising performance in the classification of SZ, it still has limitations. SZ is a clinical syndrome composed of a variety of symptoms, and the pathogenesis of patients with different symptoms may be different. The construction of an individualized accurate diagnosis model suitable for subtypes of schizophrenia is expected to further improve the diagnostic accuracy of SZ. Furthermore, SZ shows a heritability of approximately 80% [51], and studies have found that clinical abnormalities in SZ are closely correlated with the expression of associated genes [52]. Consequently, the introduction of genetic data is beneficial for strengthening the capacity of feature representation and may be applied to predict genes associated with the risk of illness. Future work should use deep learning methods to integrate genetic and imaging data for auxiliary diagnosis of SZ is the next promising research direction.

### V.  Conclusion

In this study, an effective algorithm based on a Multi-SIGATnet was proposed to classify the MRI data of SZ. This method fuses the data of fMRI and sMRI and obtains more comprehensive and rich features through information complementation between modalities, thereby improving the classification performance of the model. Furthermore, this study proposed a sparse interaction mechanism to eliminate redundant features while enhancing salient features. Edge features were injected into the Multi-SIGATnet model during the feature update phase to consider the impact of brain connections on classification performance. Compared to other methods, the Multi-SIGATnet method achieved promising results in SZ classification. Experimental analysis shows that brain-based topology modeling is crucial, and appropriate sparse interactions between brain regions can improve the performance of the model. Compared with the experimental results of existing models, the method based on graph neural network can improve the classification ability of SZ, which is expected to provide some support for computer-aided diagnosis.

ACKNOWLEDGMENT

This work was supported by the grant from National Key R&D Program of China 2018YFA0701400, Chinese medicine (ethnic medicine) frontier research and development innovation team of Sichuan Administration of Traditional Chinese Medicine( No. 2022C010), the Sichuan Provincial Program of Traditional Chinese Medicine of China (2024ZD014), and the Fundamental Research Funds for the Central Universities of China, Southwest Minzu University (ZYN2023098)